\documentclass[lettersize, apacite, twoside, HRI]{apa_HRI}
\usepackage{times}
\usepackage{natbib}
\usepackage{graphicx}  

\usepackage{epstopdf}
\usepackage{amssymb,amsmath}
\usepackage{bbm}
\usepackage[footnotesize]{caption}
\usepackage{subcaption}
\usepackage{tikz}
\usetikzlibrary{arrows}
\usepackage{booktabs}
\usepackage{xifthen}

\DeclareMathOperator*{\argmax}{arg\,max}

\newcommand{\xxnote}[3]{}
\ifx\hidenotes\undefined
  \usepackage{color}
  \renewcommand{\xxnote}[3]{\color{#2}{#1: #3}}
\fi

\newcommand{\ccw}{open triangle 45-}
\newcommand{\cw} {-open triangle 45}
\newcommand{\hdir}{\cw}
\newcommand{\rdir}{\cw}
\newcommand{\scol}{red}

\newcommand{\scenefig}[5]
{
\ifthenelse{\equal{#1}{+}} {\renewcommand{\hdir}{\ccw}}{\renewcommand{\hdir}{\cw}}
\ifthenelse{\equal{#2}{+}} {\renewcommand{\rdir}{\ccw}}{\renewcommand{\rdir}{\cw}}
\ifthenelse{\equal{#1}{#2}}{\renewcommand{\scol}{black!30!green}}{\renewcommand{\scol}{red}}

\begin{tikzpicture}
\draw [draw=white] (-1.1,-1.1) rectangle (1.1,1.1);
\begin{scope}[rotate around={{#3}:(0,0)}]
\draw [{\scol}, thick, \hdir, rotate around={45:(0,0)}] (0,-1) arc (-90: -180:1);
\ifthenelse{\equal{#2}{c}}{
\node(1) at(0,1.0) {\LARGE \textbf{C}};
}{

\ifthenelse{\equal{#2}{s}}{
\node(1) at(0,1.0) {\LARGE \textbf{S}};
}
{

\draw [{\scol}, thick, \rdir, rotate around={45:(0,0)}] (0, 1) arc (  90:   0:1  );
}
}
\filldraw[fill=\scol!20!white, draw={\scol}, thick] (-0.4,-0.6) rectangle (.4,.6);
\ifthenelse{\equal{#4}{white}}{\filldraw[fill=#4] (0,-0.6) circle (.125);}{\filldraw[draw = #4, fill=#4] (0,-0.6) circle (.125);}
\filldraw[draw = #5, fill=#5] (0,0.6) circle (.125);
\end{scope}
\end{tikzpicture}
}

\rightheader{Planning with Verbal Communication}   
\leftheader{Nikolaidis et al.}  

\title{Planning with Verbal Communication \\for Human-Robot Collaboration}

\fourauthors{Stefanos Nikolaidis}{Minae Kwon}{Jodi Forlizzi}{Siddhartha Srinivasa}

\fouraffiliations{The Robotics Institute, Carnegie Mellon University}
{Computing and Information Science, Cornell University}{Human Computer Interaction, Carnegie Mellon University}{The Robotics Institute, Carnegie Mellon University}
\abstract{ Human collaborators coordinate effectively their actions through both verbal and non-verbal communication. We believe that the the same should hold for human-robot teams. We propose a formalism that enables a robot to decide optimally between doing a task and issuing an utterance. We focus on two types of utterances: verbal commands, where the robot expresses \emph{how} it wants its human teammate to behave, and state-conveying actions, where the robot explains \emph{why} it is behaving this way. Human subject experiments show that enabling the robot to issue verbal commands is the most effective form of communicating objectives, while retaining user trust in the robot. Communicating \emph{why} information should be done judiciously, since many participants questioned the truthfulness of the robot statements.}


\begin{document}
\maketitle

\section{Introduction}
\label{sec:introduction}
The recent development of robotic systems designed to co-exist with humans highlights the need for systems that can act as trustworthy partners in a team, while collaborating effectively with their human counterparts.

\begin{figure}[t!]
\centering
\begin{subfigure}[b]{0.64\linewidth}
 \includegraphics[width=1.0\linewidth]{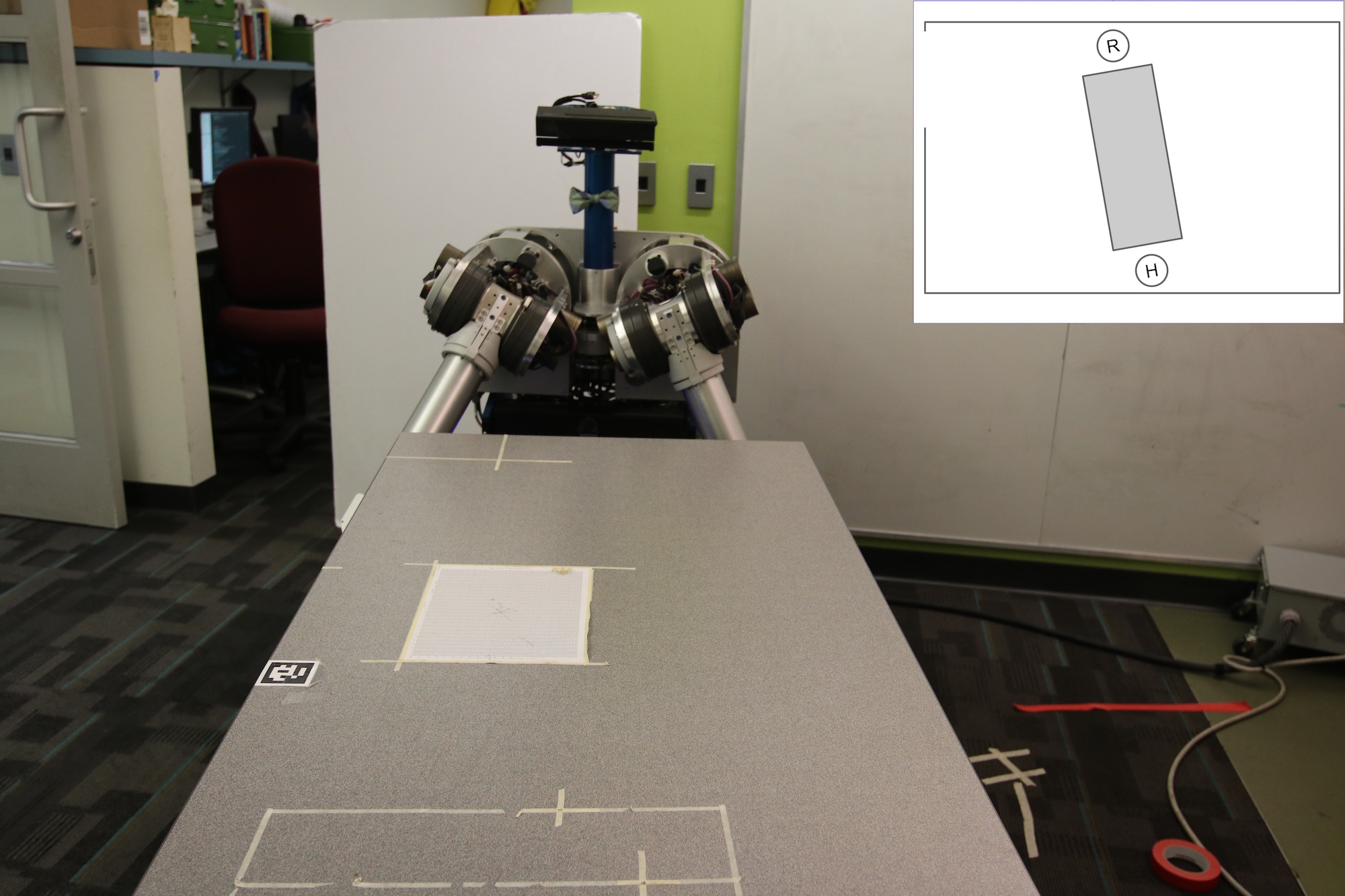}
\caption{}
 \label{fig:tablecarrying}
 \end{subfigure}
\begin{subfigure}[b]{0.318\linewidth}
 \vspace{0.2cm}
\centering
 \includegraphics[width=1.0\linewidth]{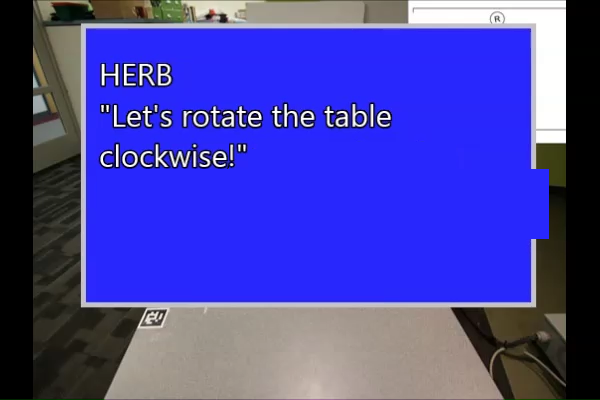}
  \includegraphics[width=1.0\linewidth]{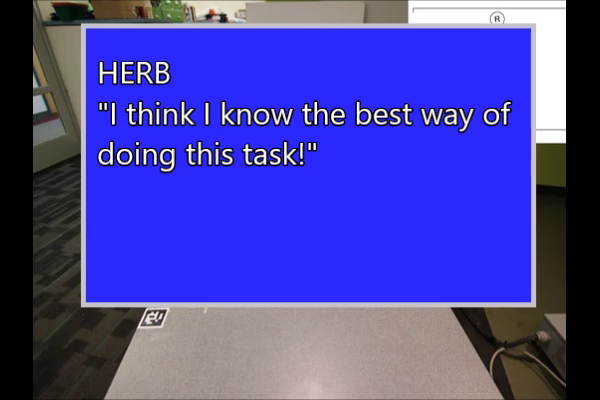}
\caption{}
\label{fig:table_robot_verbal}
\end{subfigure}
\caption{(a) Human-robot table carrying task. (b) The robot issues a verbal command. (c) The robot issues a state-conveying action.}
\label{fig:TableCarryingTask}
\end{figure}

This is often challenging; inexperienced users often form inaccurate expectations of a robot's capabilities~\citep{takayama2010}, and thus come up with suboptimal strategies to complete a collaborative task. We use as example a table carrying task, where a human-robot team works together to move a table outside of a room~(Fig.~\ref{fig:TableCarryingTask}). The pair can accomplish this in two ways. The first involves the robot facing the door and the second involves the robot facing the room. Having the robot facing the room is a suboptimal strategy, since its on-board sensors will not be able to detect the exit accurately. In the case where the human starts with this strategy, the robot should guide them towards a better way of doing the task.  

A simple approach would be to always have the robot insist on the best goal. However, our pilot studies have shown that several participants wanted to impose their preferences on the robot. If the robot did not adapt, people lost their trust in the system and were hesitant to work with the robot in the future. 
While we want to improve human-robot team performance, we do not want this to be to the detriment of user trust in the robotic system. 

In previous work,~\citet{nikolaidis2016formalizing,nikolaidis2017human} proposed a mutual-adaptation formalism, where robots can infer the adaptability of their human teammate. If their teammate is adaptable, the robot will insist on the optimal strategy. Otherwise, the robot will adapt to their strategy, in order to retain their trust. While the authors did not consider verbal communication, previous studies suggest that there is a significant benefit when teammates verbally communicate intentions and expectations~\citep{wang2013virtual,parikh2014,eccles2004,bowers1998}. 

We generalize the mutual-adaptation formalism of previous work to include verbal communication. Our generalized formalism enables a robot to~\emph{combine optimally verbal communication and actions towards task completion} to guide a human teammate towards a better way of doing a collaborative task. 

We focus on the robot verbally communicating two types of information: \emph{how} the robot wants them to behave, and \emph{why} the robot is behaving this way. Therefore, we identify two types of verbal communication: \emph{verbal commands}, where the robot asks the human to take a specific action, i.e., ``Let's rotate the table clockwise'', and \emph{state-conveying actions}, i.e., ``I think I know the best way of doing the task,'' where the robot informs the human about its internal state, which captures the information that the robot uses in its decision making (Fig.~\ref{fig:TableCarryingTask}).

We then formulate and learn from data a mixed-observability Markov decision process (MOMDP) model. The model allows the robot to reason about the human internal state, in particular about how willing the human teammate is to follow a robot task action or a robot verbal command, and to optimally choose to take a task action or issue a communication action. 

We conducted an online human subjects experiments featuring a table carrying task and compared results between three instantiations of our formalism: one that combines task actions with verbal communication, one that combines task actions with state-conveying actions, and the formalism from previous work~\citep{nikolaidis2016formalizing} that considers only non-verbal task actions, i.e., rotating the table in the table carrying example. Results show that adding verbal commands to the robot decision making is the most effective form of interaction; 100\% of participants changed their strategy towards a new, optimal goal demonstrated by the robot in the first condition. On the other hand, only 60\% of participants in the non-verbal condition adapted to the robot. Trust ratings were comparable between the two conditions. Interestingly, state-conveying actions did not have a similar positive effect, since participants did not believe that the robot was truthful. These results are encouraging, but also leave room for further investigation of different ways that people interpret robot verbal behaviors in collaborative settings.

\section{Relevant Work}
\subsection{Verbal Communication in Human Teams}
Verbal discourse is a joint activity~\citep{clark1994discourse}, where participants need to establish a shared understanding of their mutual knowledge base. This shared understanding, also called common ground, can be organized into two types: a communal common group, which represents universal shared knowledge, and personal common groups which represent mutual knowledge gathered from personal experience~\citep{clark1994discourse,clark1996communities}. People develop personal common ground by contributing new information, which enables participants in the conversation to reach a mutual belief. This belief, known as grounding~\citep{clark1989contributing}, indicates that they have understood the information as the speaker intended. \citet{grice1975logic} has shown that grounding is achieved when people avoid expending unnecessary effort to convey information.

Previous work has shown that establishing grounding through verbal communication can improve performance, even when combined with other types of feedback.~\citet{wang2013virtual} show that the efficiency of haptic communication was improved only after dyads were first given a learning period in which they could familiarize themselves with the task using verbal communication. \citet{parikh2014} find that for a more complicated task, verbal feedback coupled with haptic feedback has a significant positive effect on team performance, as opposed to haptic feedback alone.  In general, verbalization is more flexible than haptic feedback, since it allows for the communication of more abstract and complex ideas \citep{eccles2004}, while it can facilitate a shared understanding of the task~\citep{bowers1998}. 

However, verbal communication is costly in terms of time and cognitive resources~\citep{eccles2004}. For example, according to \citet{clark1991grounding}, it costs time and effort to formulate coherent utterances, especially when talking about unfamiliar objects or ideas. Receivers also experience costs in receiving and understanding a message; listening and understanding utterances can be especially costly when contextual cues are missing and the listener needs to infer the meaning. Thus, after teams have a shared understanding of the task, it may be beneficial  to switch to a less costly mode of communication, such as haptic feedback. In fact,~\citet{kuc2013} show that haptic feedback increases a perceived sense of presence and collaboration, making interaction easier. Haptic communication has been shown to be especially effective in tasks that involve deictic referencing and guiding physical objects~\citep{moll2009}.

We draw upon these insights to propose a formalism for combining verbal communication and task actions, in order to guide a human teammate towards a better way of doing a task. We investigate the effect of different types of verbal communication in team performance and trust in the robot.
\subsection{Verbal communication in Human-Robot Teams}
Verbal communication in human-robot teams has been shown to affect collaboration, as well as people's perception of the robot~\citep{mavridis2015review,thomaz2016computational, grigore2016talk}. Robot dialog systems have mostly supported human-initiated or robot-initiated communication in the form of requests. An important challenge for generating legible verbal commands has been symbol grounding~\citep{mavridis2015review,tellex2011approaching}, which is described as the ability of the robot to map a symbol to a physical object in the world.~\citet{tellex2011approaching} presented a model for inferring plans from natural language commands; inverting the model enables a robot to recover from failures, by communicating the need for help to a human partner using natural language~\citep{tellex2014asking}.~\citet{khan2009minimal} proposed a method for generating the minimal sufficient explanation that explains the policy of a Markov decision process, and \citet{wang2016impact} proposed generating explanations about the robot's confidence on its own beliefs. Recent work by~\citet{hayes2017improving} has generalized the generation of explanations of the robot policies to a variety of robot controllers.

Of particular relevance is previous work in the autonomous driving domain~\citep{koo2015did}. Messages that conveyed ``how'' information, such as ``the car is breaking,'' led to poor driving performance, whereas messages containing ``why'' information, such as ``There is an obstacle ahead,''  were preferred and improved performance. Contrary to the driving domain, in our setting the human cannot verify the truthfulness of the robot ``why'' action. Additionally, unlike driving, in a physical human-robot collaboration setting there is not a clearly right action that the robot should take, which brings the human to a state of uncertainty and disagreement with the robot. In agreement with~\citet{koo2015did}, our results show the importance of finding the right away to explain robot behavior to human teammates.

Our work is also relevant to the work by~\citet{st2015robot}. The authors explored communication in a shared-location collaborative task, using three different types of verbal feedback: self-narrative (e.g., ``I'll take care of X''), role-allocative (e.g., ``you handle X'') and empathetic (e.g., ``Oh no'' or ``Great''). They showed that feedback improves both objective and subjective metrics of team performance. In fact, the robot's \emph{verbal commands} (``Let's rotate the table clockwise'') and \emph{state-conveying actions} (``I think I know the best way of doing the task,'') of our work resemble the role-allocative and self-narrative feedback. Additionally,~\citet{oudah2015learning} integrated verbal feedback about past actions and future plans into a learning algorithm, resulting in improved human-robot team performance in two game scenarios.

Contrary to existing work~\footnote{In~\citet{devin2016implemented}, the robot reasons over the human mental state, which represents the human knowledge of the world state and of the task goals. The human mental state is assumed to be fully observable by the robot.}, our formalism enables the robot to reason about the effects of various types of verbal communication on the future actions of different human collaborators, based on their \emph{internal state}. The human internal state captures inter-individual variability. Integrating it as a latent variable in a partially observable stochastic process allows the robot to infer online the internal state of a new human collaborator and decide when it is optimal to give feedback, as well as which type of feedback to give. 



\subsection{Planning Under Uncertainty in Human-Robot Collaboration}
 In previous work, partially observable Markov decision processes (POMDP) have enabled robotic teammates to coordinate through communication~\citep{barrett2014communicating} and software agents to infer the intention of human players in game AI applications~\citep{macindoe2012pomcop}. In human-robot collaboration, the model has been successfully applied to real-world tasks, such as autonomous driving where the robot car interacts with pedestrians and human drivers~\citep{bai2015intention, bandyopadhyay2013intention,galceran2015multipolicy}.~\citet{nikolaidis2016formalizing, nikolaidis2017human} recently proposed a human-robot mutual adaptation formalism, where the robot builds online a model of how willing the human is to adapt to the robot, based on their \emph{adaptability}. The formalism models the human adaptability as a latent variable in a mixed-observability Markov decision process. This enables the robot to update its estimate on the adaptability of its human collaborator through interaction and actively guide its teammate towards a better way of doing their task. In this work, we generalize the human-robot mutual adaptation formalism by incorporating verbal communication from the robot to the human.


\section{Problem Setting}
A human-robot team can be treated as a multi-agent system, with world state $x_{w} \in X_{w}$, robot action $a_r \in A_r$, and human action $a_h \in A_h$. The system evolves according to a stochastic state transition function $T \colon X_{w} \times A_r \times A_h \rightarrow \Pi(X_{w})$.  At each time step, the human-robot team receives a real-valued reward $R(x_{w}, a_r, a_h)$. Its goal is to maximize the expected total reward over time: ${\sum_{t=0}^{\infty} {\gamma^t R(t)}}$, where the discount factor $\gamma\in [\,0,1)$ gives higher weight to immediate rewards than future ones.

The robot's goal is to compute an optimal policy $\pi_{r}^{*}$ that maximizes the expected
total discounted reward:
\begin{align}
        \pi_{r}^{*} = \underset{\pi_r}{\argmax} \ \underset{}{\mathbb{E}} \bigg{[} \underset{t=0}{\overset{\infty}{\sum}} \gamma^t \mathcal{R}\big{(} x(t), a_r(t), a_h(t)\big{)}\mid \pi_r, \pi_h  \bigg{]}
    \label{eq:MAMDP}
\end{align}

The expectation is taken over the human behavioral policies $\pi_h$ and the sequence of uncertain state transitions over time.  In order to solve the optimization problem from~Eq.\ref{eq:MAMDP}, the robot needs access to the human policies $\pi_h$. In Sec.~\ref{subsec:BAM} we provide an overview of the Bounded-Memory Adaptation Model, presented in~\cite{nikolaidis2016formalizing}, which specifies the human policies $\pi_h$.

\subsection{Bounded-Memory Adaptation Model} \label{subsec:BAM}
 BAM defines a set $M$ of \textit{modal policies} or \textit{modes} and assumes that the human switches among the modes stochastically.  A mode $m\colon X_{w}  \times A_r \times A_h \rightarrow \{0,1\}$ is a deterministic policy that maps the current world state to joint human-robot actions.  At each time step, the human follows a mode $m_h \in M$ and observes that the robot follows a mode $m_r \in M$. 

 For instance, in the table carrying task of Fig.~\ref{fig:TableCarryingTask}, one mode informally represents the human and the robot moving the table clockwise. Another mode can represent the human and the robot moving the table counter-clockwise. Intuitively, a set of modes captures the different ways that lead to task completion.

If human and robot disagree, that is they follow different modes, the human may switch from their mode $m_h$ to the robot's mode $m_r$ at the next time step. We assume that this occurs with probability~$\alpha$. If $\alpha=1$, the human switches to $m_r$ almost surely. If $\alpha=0$, the human insists on the original mode $m_h$ and does not adapt at all. Intuitively, $\alpha$ captures the human's inclination to adapt. We define $\alpha$ the human \textit{adaptability}.

When inferring the robot mode $m_r$, the human may take into account not the whole history, but only the last $k$ time-steps in the past.  This assumption of ``bounded rationality'' was first proposed by Herbert Simon: people often do not have the time and cognitive capabilities to make perfectly rational decisions~\citep{simon1979rational}.  In game theory,  bounded rationality has been modeled by assuming that players have a ``bounded memory'' or ``bounded recall" and base their decisions on recent observations~\citep{powers2005learning, monte2014learning,aumann1989cooperation}. We find this assumption particularly pertinent in a fast-paced task, such as the table carrying task, where the human collaborator has limited time to choose their actions.

\subsection{Mutual Adaptation Formalism via Task Actions} \label{subsec:MOMDP}
In this section we describe the integration of BAM in the robot decision making process using the MOMDP formulation by~\cite{nikolaidis2016formalizing}. A MOMDP uses proper factorization of the observable and unobservable state variables $S: X \times Y$, reducing the computational load~\citep{ong2010planning}. 

We include in the set of observable variables $X$ the modal policies followed in the last $k$ time-steps, so that $X:X_{w}\times M^k \times M^k$. $X_{w}$ is the finite set of task-steps that signify progress towards task completion and $M$ is the set of modal policies followed by the human and the robot in a history length $k$. The partially observable variable $y$ in this case is identical to the human adaptability $\alpha$, so that $Y \equiv \mathcal{A}$.~\citet{nikolaidis2016formalizing} consider $\alpha$ to be fixed throughout the task. We define $\mathcal{T}_{x}$ the transition function that specifies how the observable state changes given a human and a robot action. We denote as $\pi_h$ the stochastic human policy. The latter gives the probability of a human action $a_h$ at state $s$, based on the BAM human adaptation model. 
The belief update for the MOMDP in this model is: 
\begin{equation}
\label{e:beliefupdate}
\begin{split}
b'(\alpha') = &\eta \sum_{\alpha \in \mathcal{A}}  \sum_{a_h \in A_h} \mathcal{T}_{x}(x, y, a_{r}, a_{h}, x')\pi_h(s, a_h) b(y)
\end{split}
\end{equation}

We can then solve the MOMDP for a robot policy $\pi^{*}_r$ (Eq.~\ref{eq:MAMDP}) that takes into account the robot belief on the human adaptability, while maximizing the agent's expected total reward. Solving exactly the MOMDP is intractable; we find an approximate solution using the SARSOP solver~\citep{kurniawati2008sarsop}, a point-based approximation algorithm which, combined with the MOMDP formulation, can scale up to hundreds of thousands of states~\citep{bandyopadhyay2013intention}. 

The policy execution is performed online in real time and consists of two steps. First, the robot uses the current belief to select the action $a_r$ specified by the policy. Second, it uses the observation of the new state to update the belief on $\alpha$ (Eq.~\ref{e:beliefupdate}). 

The presented human-robot mutual adaptation formalism allows the robot to guide the human towards a better way of doing the task via disagreement through actions. In Sec.~\ref{sec:VerbalCommunication}, we generalize the proposed model, allowing for the robot to communicate with the human through verbal utterances, as well.

\section{Planning with Verbal Communication} \label{sec:VerbalCommunication}

We identify two types of verbal communication: \emph{verbal commands}, where the robot asks the human to take a specific action, i.e., ``Let's rotate the table clockwise'', and \emph{state-conveying actions}, i.e., ``I think I know the best way of doing the task,'' where the robot informs the human about its internal state.

\subsection{Robot Verbal Commands} \label{subsec:RobotVerbalCommands}
We define as verbal command a robot action, where the robot asks the human partner to follow an action $a_h \in A_h$ specified by some mode $m_r \in M$. We use the notation $a^{w}_r \in A^{w}_r$ for robot task actions that affect the world state and $a^{c}_r \in A^{c}_r$ for robot actions that correspond to the robot giving a verbal command to the human. We assume a known bijective function $f: A^{h} \rightarrow A^{c}_r$ that specifies an one-to-one mapping of the set of human actions to the set of robot commands. 


\noindent\textbf{Human Compliance Model.}
Given a robot command $a^c_r \in A^c_r$, the human can either ignore the command and insist on their mode $m_h \in M$, or switch to a mode $m_r \in M$ inferred by $a^c_r$ and take an action $a_h \in A_h$ specified by that mode. We assume that this will happen with probability $c$, which indicates the human \textit{compliance} to following robot verbal commands. We model human compliance separately to human \textit{adaptability}, drawing upon insights from previous work on verbal and non-verbal communication which shows that team behaviors can vary in different interaction modalities~\citep{wang2016,chellali2012}.

\noindent\textbf{MOMDP Formulation.}
We augment the formulation of Section~\ref{subsec:MOMDP}, to account for robot verbal commands, in addition to task actions: the set of robot actions $A_r$ is now $A_r : A^{w}_r \times A^{c}_r$. 

The set of observable variables $X$ includes the modal policies followed in the last $k$ time-steps, so that $X:X_{w}\times M^k \times M^k \times B$. Compared to the formulation of Sec.~\ref{subsec:MOMDP}, we additionally include a flag $B \in \{0,1\}$, that indicates whether the last robot action was a verbal command or a task action. 
The set of partially observable variables includes both human \textit{adaptability} $\alpha$ in $\mathcal{A}$ and \textit{compliance} $c \in \mathcal{C}$, so that $Y: \mathcal{A} \times \mathcal{C}$. We assume both $\alpha$ and $c$ to be fixed throughout the task.

The belief update for the MOMDP in this model is: 
\begin{equation}
\label{e:beliefupdate-verbal}
\begin{split}
b'(\alpha', c') = &\eta \sum_{\alpha \in \mathcal{A}} \sum_{c \in \mathcal{C}} \sum_{a_h \in A_h} \mathcal{T}_{x}(x, y, a_{r}, a_{h}, x') \pi_h(x, \alpha, c, \ a_h) b(\alpha, c)
\end{split}
\end{equation} 

\begin{figure}[t!]
 \centering
  \includegraphics[width=0.35\linewidth]{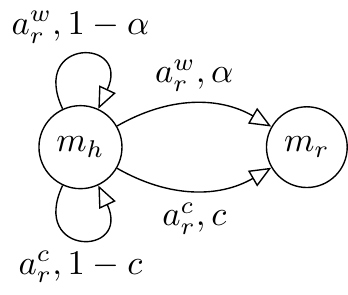}
  \caption{Human adaptation model that accounts for verbal commands. If the robot gave a verbal command $a^c_r$ in the previous time-step, the human will switch modes with probability $c$. Instead, if the robot took an action $a^w_r$ that changes the world state, the human will switch modes with probability $\alpha$.}
  \label{fig:HumanAdaptationComplex}
 \end{figure}

The human policy $\pi_h(x, \alpha, c, \ a_h)$ captures the probability of the human taking an action $a_h$ based on their adaptability and compliance. In particular, if $B \equiv 1$, indicating that the robot gave a verbal command in the last time-step, the human will switch to a mode $m_r \in M$ specified by the previous robot command $a^c_r$  with probability $c$, or insist on their human mode of the previous time-step $m_h$ with probability $1-c$. If $B \equiv 0$, the human will switch to a mode $m_r \in M$ specified by the robot action $a^w_r$  with probability $\alpha$, or insist on their human mode of the previous time-step $m_h$ with probability $1-\alpha$. Fig.~\ref{fig:HumanAdaptationComplex} illustrates the model of human decision making that accounts for verbal commands. 

As in Sec.~\ref{subsec:MOMDP}, we then solve the MOMDP for a robot policy $\pi^{*}_r$ (Eq.~\ref{eq:MAMDP}). This time, the robot optimal policy will take into account both \emph{the robot belief on human adaptability and the robot belief on human compliance}. It will decide optimally, based on this belief, whether to take a task action or issue a verbal command. We show that this improves the adaptation of human teammates in Sec.~\ref{sec:Evaluation}.

\subsection{Communication of Robot Internal State}  \label{subsec:RobotInternalState}

Previous work~\citep{van2011team} has shown that communicating internal states among team members allows participants to form \textit{shared mental models}. Empirical evidence suggests that mental model similarity improves coordination processes which, in turn, enhance team performance~\citep{mathieu2000influence, Marks_2002}. The literature presents various definitions for the concept of ``shared mental models'' \citep{LanganFoxCL00}. ~\citet{Marks_2002} state that mental models represent ``the content and organization of inter-role knowledge held by team members within a performance setting.'' According to \citet{Mathieu_2000}, mental models are ``mechanisms whereby humans generate descriptions of system purpose and form, explanations of system functioning and observed system states and prediction of future system states \ldots\ and they help people to describe, explain and predict events in their environment.'' Other work~\citep{goodrich2013toward, kiesler2002mental, nikolaidis2013} has shown the effect of shared mental models on team performance for human-robot teams, as well. Using these insights, we propose a way for the robot to communicate its internal state to the human.

\noindent\textbf{State Conveying Actions.} 
We define as state-conveying action a robot action, where the robot provides to the human information about its decision making mechanism. We define a set of state-conveying actions $a^s_r \in A^s_r$. These actions do not provide information about the robot mode, but we expect them to increase the human \textit{adaptability} and \textit{compliance} levels. In autonomous driving, users showed greater system acceptance, when the system explained the reason for its actions~\citep{koo2015did}.

\noindent\textbf{MOMDP Formulation.}
We describe the integration of state-conveying actions in the MOMDP formulation. 

The set of robot actions includes task-based actions and state-conveying actions, so that:
$A_r : A^{w}_r \times A^{s}_r$.
We model an action $a^s_r$ as inducing a stochastic transition from a human adaptability $\alpha \in \mathcal{A}$ to $\alpha' \in \mathcal{A}$, and $c \in \mathcal{C}$ to $c' \in \mathcal{C}$. Formally, we define the transition functions for the partially observable variables $\alpha$, so that: $\mathcal{T}_{\alpha}: \mathcal{A} \times A^s_r \rightarrow \Pi(\mathcal{A})$ and  $\mathcal{T}_{c}: \mathcal{A} \times A^s_r \rightarrow \Pi(\mathcal{C})$. We note that the task actions $a_r \notin A^{s}_r$ do not change $\alpha$  and $c$. 

The belief update now becomes: 
\begin{equation}
\label{e:beliefupdate-sc}
\begin{split}
b'(\alpha',c') = &\eta \sum_{\alpha \in \mathcal{A},~c \in \mathcal{C}}\mathcal{T}_{\alpha}(\alpha,a_{r}, \alpha') \mathcal{T}_{c}(c,a_{r}, c') \sum_{a_h \in A_h}\mathcal{T}_{x}(x, y, a_{r}, a_{h}, x') \pi_h(x, \alpha, c, a_h) b(\alpha,c)
\end{split}
\end{equation}

As in Sec.~\ref{subsec:MOMDP}, we solve the MOMDP for a robot policy $\pi^{*}_r$ (Eq.~\ref{eq:MAMDP}). The robot policy will decide optimally whether to take a task action or a state-conveying action. Intuitively, if the inferred human adaptability  / compliance is low, the robot should take a state-conveying action to make the human teammate more adaptable / compliant. Otherwise, it should take a task action, expecting the human to adapt / follow a verbal command. We examine the robot behavior in this case in Sec.~\ref{sec:Evaluation}.

\section{Model Learning} \label{sec:model-learning}
To compute the belief update of Eq.~\ref{e:beliefupdate},~\ref{e:beliefupdate-verbal} and ~\ref{e:beliefupdate-sc}, we need a prior distribution~\footnote{We are using the term prior distribution and prior belief interchangeably.} over the human adaptability and compliance values. We additionally need to specify the $\mathcal{T}_{\alpha}$ and $\mathcal{T}_{c}$ that indicate how the adaptability and compliance will change, when the robot takes a state-conveying action. 

In previous work,~\citet{nikolaidis2016formalizing} assumed a uniform prior on human adaptability. While we could do the same in this work, this would ignore the fact that people may in general have different \emph{a priori} dispositions towards adapting to the robot when it takes a task action and towards following a robot verbal command. In fact,~\citet{albrecht2015empirical} have empirically shown that prior beliefs can have a significant impact on the performance of utility-based algorithms. Therefore, in this section we propose a method for learning a prior distribution on human adaptability and compliance from data.

We additionally propose a method for computing the state transition function $\mathcal{T}_{\alpha}$ in Eq.~\ref{e:beliefupdate-sc}. We can use exactly the same process to compute  $\mathcal{T}_{c}$ in Eq.~\ref{e:beliefupdate-sc}, and we leave this for future work. 

\subsection{Learning Prior Distributions on Adaptability and Compliance} \label{subsec:LearningPriorBeliefs}

When integrating compliance and adaptability, we hypothesize that users are \textit{a priori} more likely to change their actions after a robot issues a verbal command, compared with the robot taking a different task action. To account for this, we compute a probability distribution over human adaptability and compliance, which the robot will use as \textit{prior} in the belief update of the MOMDP formulation.


\begin{figure}[t!]
\centering
\begin{subfigure}[b]{0.33\linewidth}
\includegraphics[width=\linewidth]{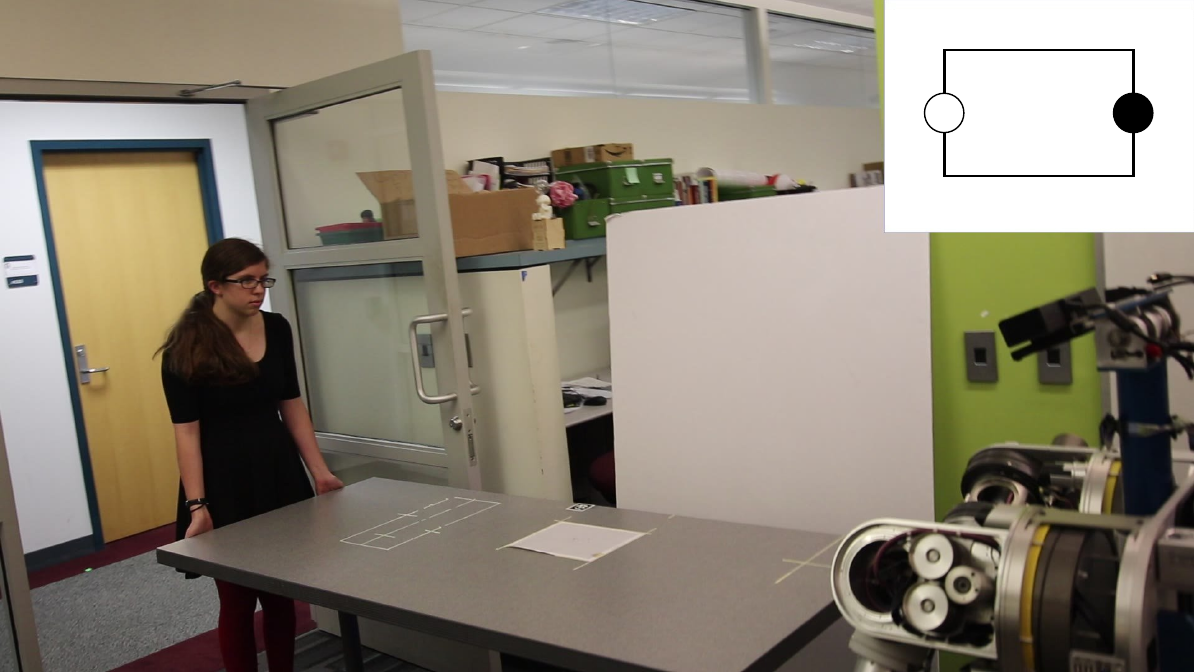}
\includegraphics[width=\linewidth]{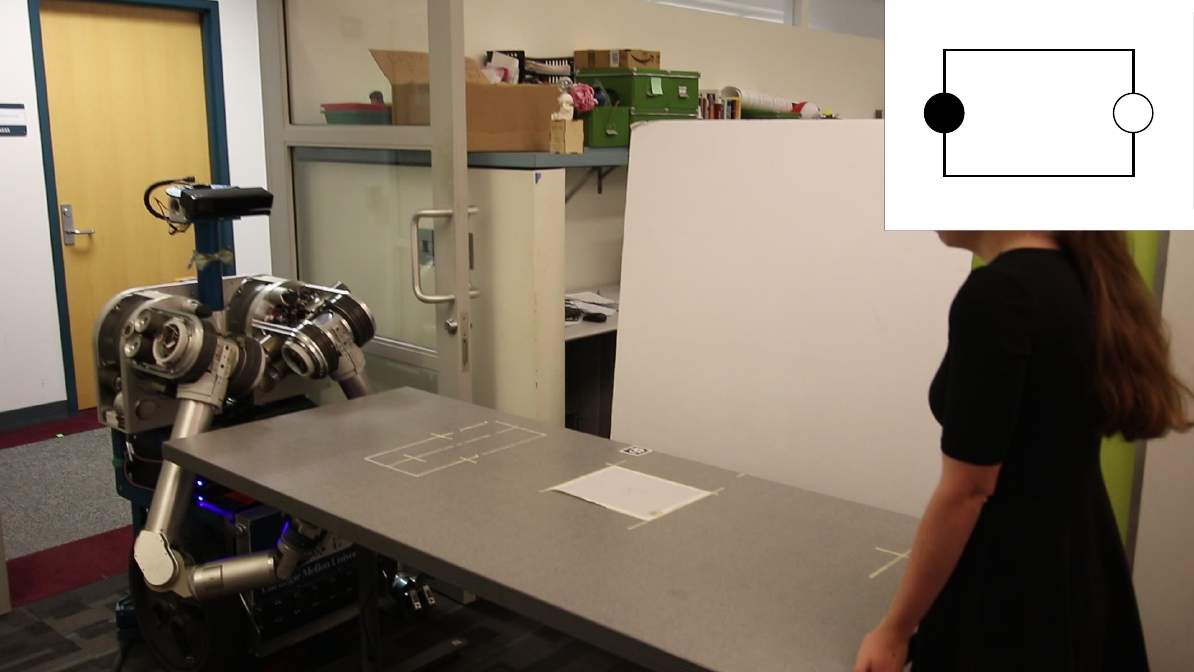}
\caption{}
 \label{fig:goals}
 \end{subfigure}
\begin{subfigure}[b]{0.4\linewidth}
\centering
\includegraphics[width=\linewidth]{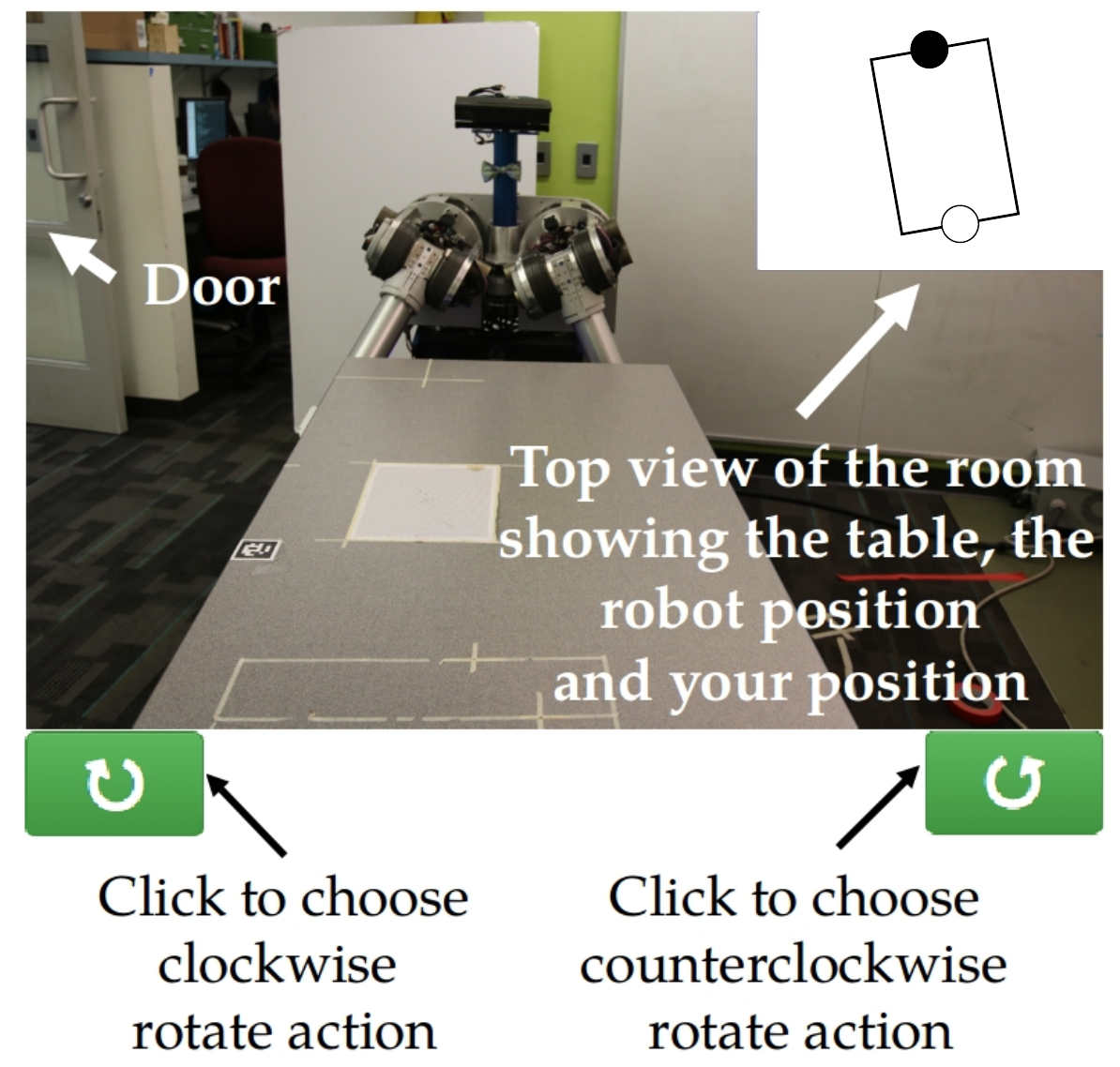}
\caption{}
\label{fig:UserInterface}
\end{subfigure}
\caption{(a) Rotating the table so that the robot is facing the door (top, Goal 1) is better than the other direction (bottom, Goal 2), since the exit is included in the robot's field of view and the robot can avoid collisions. (b) UI with instructions.}
\label{fig:TableCarryingTask2}
\end{figure}

\noindent\textbf{Data Collection Setup}. 
To collect data, we used the table carrying task setting by~\citet{nikolaidis2016formalizing}. In this task, which is performed online via video playback, human and HERB~\citep{srinivasa2010herb}, an autonomous mobile manipulator, must work together to carry a table out of the room.  There are two strategies: the robot facing the door (Goal~1) or the robot facing away from the door (Goal~2). We assume that Goal~1 is the optimal goal, since the robot's forward-facing sensor has a clear view of the door, resulting in better overall task performance. Not aware of this, an inexperienced human partner may prefer Goal~2. In our computational model, there are two modes; one with rotation actions towards Goal 1, and one with rotation actions towards Goal 2. Disagreement occurs when human and robot attempt to rotate the table towards opposite directions. We first instructed participants in the task and asked them to choose one of the two goal configurations (Fig.~\ref{fig:TableCarryingTask2}), as their preferred way of accomplishing the task. To prompt users to prefer the sub-optimal goal, we informed them about the starting state of the task, where the table was slightly rotated in the counter-clockwise direction, making the sub-optimal Goal~2 appear closer. Once the task started, the user chose the rotation actions by clicking on buttons on a user interface (Fig.~\ref{fig:UserInterface}). All participants executed the task twice.

\noindent\textbf{Manipulated Variables}.
We manipulated the way the robot reacted to the human actions. When the human chose a rotation action towards the sub-optimal goal, the table did not move and in the first condition a message appeared on the screen notifying the user that they tried to rotate the table in a different direction then the robot. In the second condition, the robot was illustrated as speaking to the user, prompting them to move the table towards the opposite direction (Figure~\ref{fig:table_robot_verbal})-top. In both conditions, when the user moved the table towards the optimal goal, a video played showing the table rotating.

\noindent\textbf{Learning Prior Beliefs.} 
\textit{Adaptability}: In Sec.~\ref{subsec:BAM}, we defined as \textit{adaptability} $\alpha$ of an individual, the probability of switching from the human mode $m_h$ to the robot mode $m_r$. Therefore, we used the data from the first condition to estimate the adaptability $\hat{\alpha}_u$ for each user $u$, as the number of times the user switched modes, divided by the number of disagreements with the robot. 

\begin{equation}
\hat{\alpha}_u = \frac{\# \text{times user $u$ switched from $m_h$ to $m_r$}}{\# \text{disagreements}}
\label{eq:alpha_u}
\end{equation}

Intuitively, a very adaptable human will switch from $m_h$ to $m_r$ after only one disagreement with the robot. On the other hand, a non-adaptable human will insist and disagree with the robot a large number of times, before finally following the robot goal. 

\textit{Compliance}: In Sec.~\ref{subsec:RobotVerbalCommands}, we defined the \textit{compliance} $c$ as the probability of following a robot verbal command and switching to a robot mode $m_r \in M$.  Therefore, similarly to Eq.~\ref{eq:alpha_u}, we estimate the compliance for each user $u$ from the second condition $\hat{c}$ as follows:

\begin{equation}
\hat{c}_u = \frac{\# \text{times user $u$ switched from $m_h$ to $m_r$}}{\# \text{verbal commands}}
\label{e:c_u}
\end{equation}

\begin{figure}[t!]
\centering
  \begin{tabular}{cc}
\begin{subfigure}[l]{0.5\linewidth}
 \includegraphics[width=0.8\linewidth]{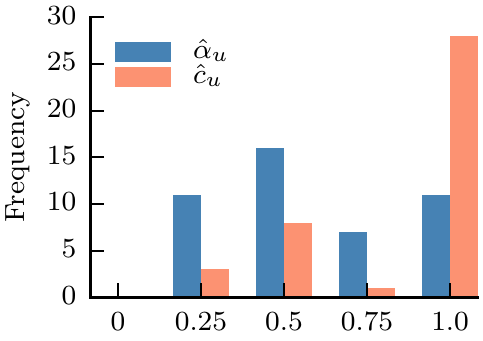}
 \caption{}
  \label{fig:hist}
   \end{subfigure}

   \begin{subfigure}[r]{0.3\linewidth}
  \includegraphics[width=0.97\linewidth]{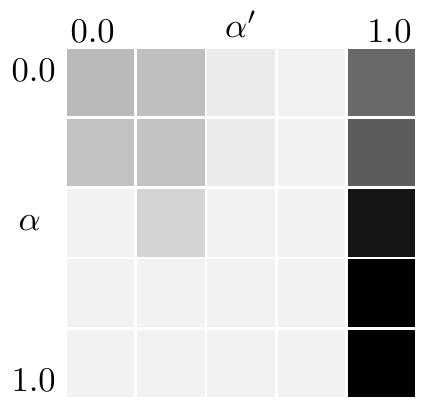}
  \caption{}
   \label{fig:state_conveying}
   \end{subfigure}
\end{tabular}
 \caption{(a) Histograms of user adaptabilities $\hat{\alpha}_u$ and compliances $\hat{c}_u$. (b) Transition matrix $\mathcal{T}_{\alpha}(\alpha,a_r^s,\alpha')$ given a robot state-conveying action $a^s_r$. Darker colors indicate higher probabilities. } 
\end{figure}

We then assume a discrete set of values for $\alpha$ and $c$, so that $\alpha \in \{0, 0.25, 0.5, 0.75, 1.0\}$ and $c \in \{0, 0.25, 0.5, 0.75, 1.0\}$, and we compute the histogram of user adaptabilities and compliances (Fig.~\ref{fig:hist}). We then normalize the histogram to get a probability distribution over user adaptabilities and a probability distribution over compliances. We use these distributions as \emph{prior beliefs} for the MOMDP model.

\noindent\textbf{Discussion.} Fig.~\ref{fig:hist} shows that most of the users adapted to the robot immediately when the robot issued a verbal command. This indicates that users are generally more likely to follow a robot verbal command than adapt to the robot through action disagreement.  

\subsection{Learning Transition Function Parameters} \label{subsec:learningtransition}
Additionally, in order to compute the belief update of Eq.~\ref{e:beliefupdate-sc}, we need to compute the state-transition function $\mathcal{T}_{\alpha}$ that represents how a state-conveying action affects the human adaptability $\alpha$. As in Sec.~\ref{subsec:LearningPriorBeliefs}, we assume $\alpha \in \mathcal{A}$, where $\mathcal{A} \in \{0,0.25,0.5,0.75,1.0\}$.

\noindent\textbf{Data Collection Setup}. We use the same table carrying setup, as in Sec.~\ref{subsec:LearningPriorBeliefs}. In the first round, participants interact with the robot executing the MOMDP policy of Sec.~\ref{subsec:MOMDP}, without any verbal communication. In the second round, we set the robot policy to move towards a goal different than the goal reached in the end of the previous round, and we have the robot take a state-conveying action in the first time-step (Fig.~\ref{fig:table_robot_verbal})-bottom. 

\noindent\textbf{Transition Function Estimation}.
Using the human and robot actions taken in the first round, we estimate the adaptability $\hat{\alpha}_u\in \mathcal{A}$ of each user $u$ using Eq.~\ref{eq:alpha_u}, rounded to the closest discrete value. We then similarly estimate the new adaptability for the same user $\hat{\alpha}'_u\in \mathcal{A}$ from the human and robot actions in the second round, after the user has observed the robot state-conveying action. We can compute the Maximum Likelihood Estimate of the transition function $\mathcal{T}_{\alpha}(\alpha, a^s_r, \alpha')$ in Eq.~\ref{e:beliefupdate-sc} from the frequency count of users that had $\alpha$, as estimated in the first round, and $\alpha'$ in the second round. Since we had only one user with $\hat{\alpha}_u \equiv 0.75$, we included the counts of adjacent entries, so that: 

\begin{equation}
\mathcal{T}_{\alpha}(\alpha, a^s_r, \alpha') = \frac{\sum_u \mathbbm{1}_{[\alpha-\delta,\alpha+\delta]}(\hat{\alpha}_u)\mathbbm{1}_{\{\alpha'\}}(\hat{\alpha}'_u)}{\sum_u \mathbbm{1}_{[\alpha-\delta,\alpha+\delta]}(\hat{\alpha}_u)}
\label{e:c_u}
\end{equation}

where $\delta=0.25$ and $\mathbbm{1}$ an indicator function.

\noindent\textbf{Discussion.} Fig.~\ref{fig:state_conveying} shows that users with intermediate or high adaptability values ($\alpha \geq 0.5$) became very adaptable ($\alpha' = 1.0$), after the robot took a state-conveying action. On the other hand, some users with low adaptability remained non-adaptable, even after the robot stated that ``[it knew] the best way of doing the task''. We investigate this effect further in Sec.~\ref{sec:Evaluation}.

\section{Evaluation} \label{sec:Evaluation}
We first simulate and comment on the different MOMDP policies using the table carrying setup of Sec.~\ref{subsec:LearningPriorBeliefs}. We then evaluate these policies in a human subject experiment.

\subsection{Simulation}

 We define the reward function in Eq.~\ref{eq:MAMDP}, so that $R_{opt} = 20$ is the reward for the optimal goal (Goal 1),  $R_{subopt} = 15$ the reward of the suboptimal goal (Goal 2), and we have $R_{other} = 0$ for the rest of the state-space. We additionally assign a discount factor of $\gamma = 0.9$. We use the MOMDP formulations of Sec.~\ref{subsec:MOMDP}, Sec.~\ref{subsec:RobotVerbalCommands} and Sec.~\ref{subsec:RobotInternalState}, and for each formulation we compute the optimal policy using the SARSOP algorithm~\citep{kurniawati2008sarsop}, which is computationally efficient and has been previously used in various robotic tasks~\citep{ bandyopadhyay2013intention}. For the policy computation, we use as prior beliefs the learned distributions from Sec.~\ref{subsec:LearningPriorBeliefs}, and as transition function $\mathcal{T}_{\alpha}$ its learned estimate from Sec.~\ref{subsec:learningtransition}.

\begin{figure*}[hp]
\setlength\tabcolsep{1.5pt}
\centering
\begin{tabular}{ccccccc}
\begin{subfigure}[l]{.08\linewidth}
\centering
\small User 1 \\ $\alpha = 1.0$ \\ $c = 1.0$ \\
\end{subfigure}
&
\begin{subfigure}[b]{.15\linewidth}
\centering
  \begin{tabular}{cc}
 \includegraphics[width=0.7\linewidth]{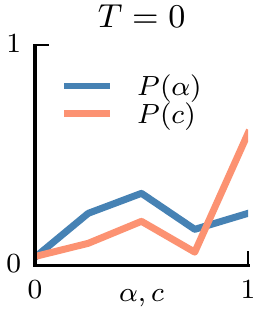}
\\
 \resizebox{0.5\linewidth}{!}{
  \scenefig{+}{-}{10}{white}{black}
   }\\
   \end{tabular}
 \label{fig:plotT0}
\end{subfigure}
&
\begin{subfigure}[b]{.15\linewidth}
\centering
  \begin{tabular}{cc}
 \includegraphics[width=0.7\linewidth]{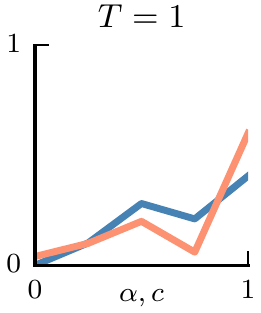}\\
  \resizebox{0.5\linewidth}{!}{
  \scenefig{-}{-}{-10}{white}{black}
   }\\
   \end{tabular}
 \label{fig:plotT0}
\end{subfigure}
\begin{subfigure}[b]{.15\linewidth}
\centering
  \begin{tabular}{cc}
 \includegraphics[width=0.7\linewidth]{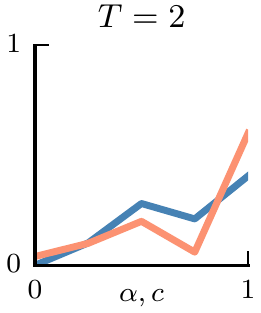}\\
    \resizebox{0.5\linewidth}{!}{
  \scenefig{-}{-}{-30}{white}{black}
   }\\  
   \end{tabular}
 \label{fig:plotT0}
\end{subfigure}
&
\begin{subfigure}[b]{.15\linewidth}
\centering
  \begin{tabular}{cc}
 \includegraphics[width=0.7\linewidth]{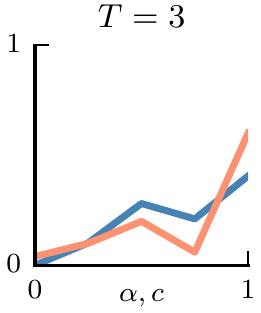}\\
    \resizebox{0.5\linewidth}{!}{
  \scenefig{-}{-}{-50}{white}{black}
   }\\ 
   \end{tabular}
 \label{fig:plotT0}
\end{subfigure}
\begin{subfigure}[b]{.15\linewidth}
\centering
  \begin{tabular}{cc}
 \includegraphics[width=0.7\linewidth]{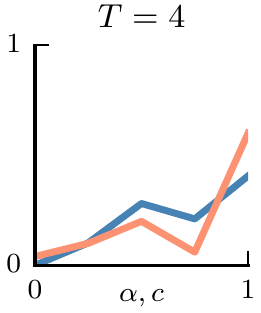}\\
    \resizebox{0.5\linewidth}{!}{
  \scenefig{-}{-}{-70}{white}{black}
   }\\ 
   \end{tabular}
 \label{fig:plotT0}
\end{subfigure}
&
\begin{subfigure}[b]{.15\linewidth}
\centering
  \begin{tabular}{cc}
 \includegraphics[width=0.7\linewidth]{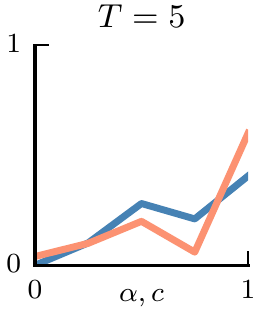}\\
  \resizebox{0.5\linewidth}{!}{
  \scenefig{-}{-}{-90}{white}{black}
   }\\ 
   \end{tabular}
 \label{fig:plotT0}
\end{subfigure}
\end{tabular}

\begin{tabular}{ccccccc}
\begin{subfigure}[l]{.08\linewidth}
\centering
\small User 2 \\ $\alpha = 0.0$ \\ $c = 0.0$ \\
\end{subfigure}
&
\begin{subfigure}[b]{.15\linewidth}
\centering
  \begin{tabular}{cc}
 \includegraphics[width=0.7\linewidth]{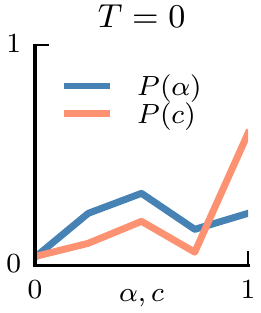}
\\
 \resizebox{0.5\linewidth}{!}{
  \scenefig{+}{-}{10}{white}{black}
   }\\
   \end{tabular}
 \label{fig:plotT0}
\end{subfigure}
&
\begin{subfigure}[b]{.15\linewidth}
\centering
  \begin{tabular}{cc}
 \includegraphics[width=0.7\linewidth]{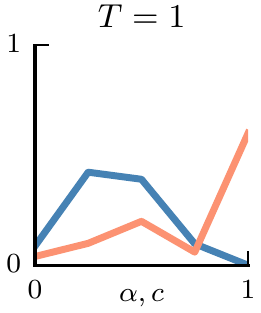}\\
  \resizebox{0.5\linewidth}{!}{
  \scenefig{+}{-}{10}{white}{black}
   }\\
   \end{tabular}
 \label{fig:plotT0}
\end{subfigure}
\begin{subfigure}[b]{.15\linewidth}
\centering
  \begin{tabular}{cc}
 \includegraphics[width=0.7\linewidth]{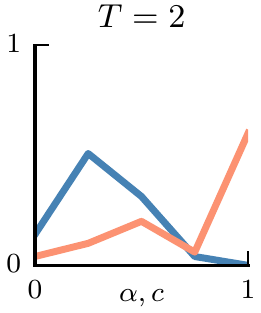}\\
    \resizebox{0.5\linewidth}{!}{
  \scenefig{+}{c}{10}{white}{black}
   }\\  
   \end{tabular}
 \label{fig:plotT0}
\end{subfigure}
&
\begin{subfigure}[b]{.15\linewidth}
\centering
  \begin{tabular}{cc}
 \includegraphics[width=0.7\linewidth]{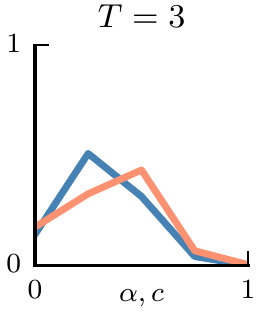}\\
    \resizebox{0.5\linewidth}{!}{
  \scenefig{+}{-}{10}{white}{black}
   }\\ 
   \end{tabular}
 \label{fig:plotT0}
\end{subfigure}
\begin{subfigure}[b]{.15\linewidth}
\centering
  \begin{tabular}{cc}
 \includegraphics[width=0.7\linewidth]{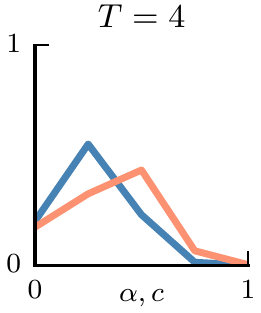}\\
    \resizebox{0.5\linewidth}{!}{
  \scenefig{+}{+}{30}{white}{black}
   }\\ 
   \end{tabular}
 \label{fig:plotT0}
\end{subfigure}
&
\begin{subfigure}[b]{.15\linewidth}
\centering
  \begin{tabular}{cc}
 \includegraphics[width=0.7\linewidth]{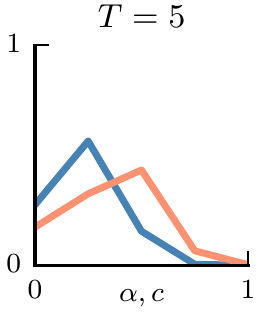}\\
  \resizebox{0.5\linewidth}{!}{
  \scenefig{+}{+}{50}{white}{black}
   }\\ 
   \end{tabular}
 \label{fig:plotT0}
\end{subfigure}
\end{tabular}
\begin{tabular}{ccccccc}
\begin{subfigure}[l]{.08\linewidth}
\centering
\small User 3 \\ $\alpha = 0.0$ \\ $c = 1.0$ \\
\end{subfigure}
&
\begin{subfigure}[b]{.15\linewidth}
\centering
  \begin{tabular}{cc}
 \includegraphics[width=0.7\linewidth]{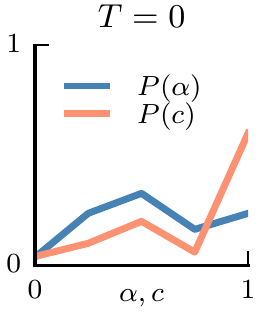}
\\
 \resizebox{0.5\linewidth}{!}{
  \scenefig{+}{-}{10}{white}{black}
   }\\
   \end{tabular}
 \label{fig:plotT0}
\end{subfigure}
&
\begin{subfigure}[b]{.15\linewidth}
\centering
  \begin{tabular}{cc}
 \includegraphics[width=0.7\linewidth]{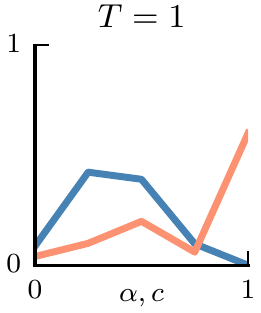}\\
  \resizebox{0.5\linewidth}{!}{
  \scenefig{+}{-}{10}{white}{black}
   }\\
   \end{tabular}
 \label{fig:plotT0}
\end{subfigure}
\begin{subfigure}[b]{.15\linewidth}
\centering
  \begin{tabular}{cc}
 \includegraphics[width=0.7\linewidth]{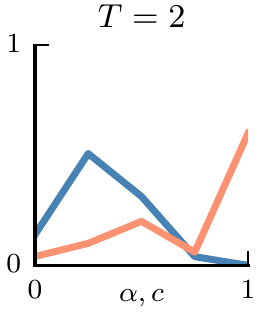}\\
    \resizebox{0.5\linewidth}{!}{
  \scenefig{+}{c}{10}{white}{black}
   }\\  
   \end{tabular}
 \label{fig:plotT0}
\end{subfigure}
&
\begin{subfigure}[b]{.15\linewidth}
\centering
  \begin{tabular}{cc}
 \includegraphics[width=0.7\linewidth]{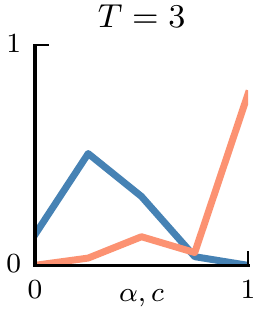}\\
    \resizebox{0.5\linewidth}{!}{
  \scenefig{-}{-}{-10}{white}{black}
   }\\ 
   \end{tabular}
 \label{fig:plotT0}
\end{subfigure}
\begin{subfigure}[b]{.15\linewidth}
\centering
  \begin{tabular}{cc}
 \includegraphics[width=0.7\linewidth]{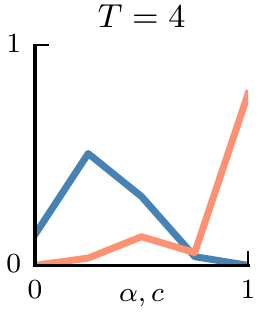}\\
    \resizebox{0.5\linewidth}{!}{
  \scenefig{-}{-}{-30}{white}{black}
   }\\ 
   \end{tabular}
 \label{fig:plotT0}
\end{subfigure}
&
\begin{subfigure}[b]{.15\linewidth}
\centering
  \begin{tabular}{cc}
 \includegraphics[width=0.7\linewidth]{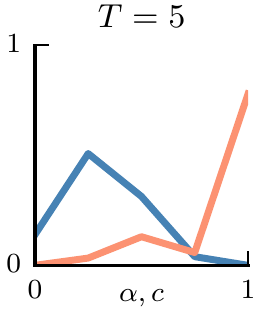}\\
  \resizebox{0.5\linewidth}{!}{
  \scenefig{-}{-}{-50}{white}{black}
   }\\ 
   \end{tabular}
 \label{fig:plotT0}
\end{subfigure}
\end{tabular}

\begin{tabular}{ccccccc}
\begin{subfigure}[l]{.08\linewidth}
\centering
\small User 4 \\  $\alpha = 0.0$ \\ (State-Conveying) \\
\end{subfigure}
&
\begin{subfigure}[b]{.15\linewidth}
\centering
  \begin{tabular}{cc}
 \includegraphics[width=0.7\linewidth]{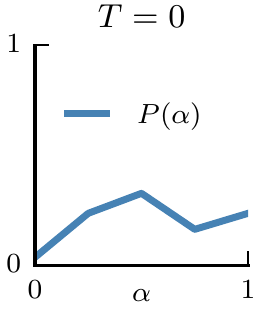}
\\
 \resizebox{0.5\linewidth}{!}{
  \scenefig{+}{-}{10}{white}{black}
   }\\
   \end{tabular}
 \label{fig:plotT0}
\end{subfigure}
&
\begin{subfigure}[b]{.15\linewidth}
\centering
  \begin{tabular}{cc}
 \includegraphics[width=0.7\linewidth]{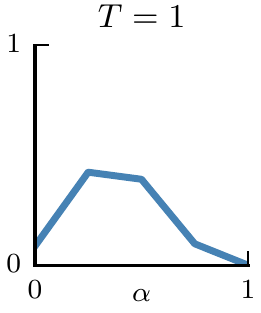}\\
  \resizebox{0.5\linewidth}{!}{
  \scenefig{+}{-}{10}{white}{black}
   }\\
   \end{tabular}
 \label{fig:plotT0}
\end{subfigure}
\begin{subfigure}[b]{.15\linewidth}
\centering
  \begin{tabular}{cc}
 \includegraphics[width=0.7\linewidth]{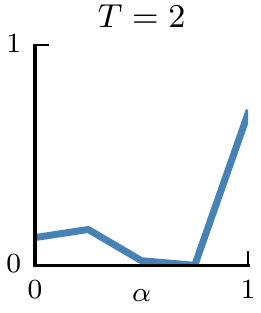}\\
    \resizebox{0.5\linewidth}{!}{
  \scenefig{+}{s}{10}{white}{black}
   }\\  
   \end{tabular}
 \label{fig:plotT0}
\end{subfigure}
&
\begin{subfigure}[b]{.15\linewidth}
\centering
  \begin{tabular}{cc}
 \includegraphics[width=0.7\linewidth]{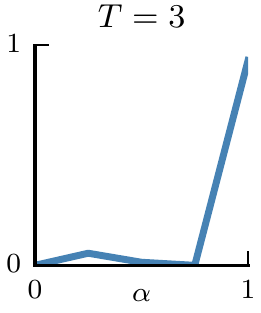}\\
    \resizebox{0.5\linewidth}{!}{
  \scenefig{-}{-}{-10}{white}{black}
   }\\ 
   \end{tabular}
 \label{fig:plotT0}
\end{subfigure}
\begin{subfigure}[b]{.15\linewidth}
\centering
  \begin{tabular}{cc}
 \includegraphics[width=0.7\linewidth]{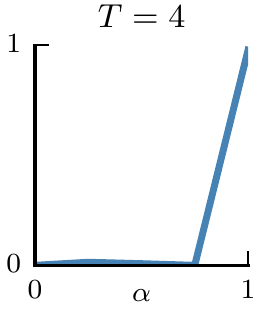}\\
    \resizebox{0.5\linewidth}{!}{
  \scenefig{-}{-}{-30}{white}{black}
   }\\ 
   \end{tabular}
 \label{fig:plotT0}
\end{subfigure}
&
\begin{subfigure}[b]{.15\linewidth}
\centering
  \begin{tabular}{cc}
 \includegraphics[width=0.7\linewidth]{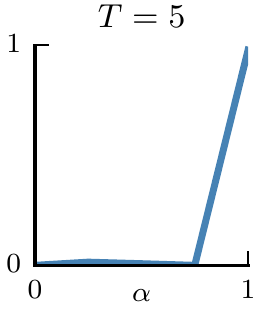}\\
  \resizebox{0.5\linewidth}{!}{
  \scenefig{-}{-}{-50}{white}{black}
   }\\ 
   \end{tabular}
 \label{fig:plotT0}
\end{subfigure}
\end{tabular}

\begin{tabular}{ccccccc}
\begin{subfigure}[l]{.08\linewidth}
\centering
\small User 5 \\  $\alpha = 0.0$ \\ (Baseline) \\
\end{subfigure}
&
\begin{subfigure}[b]{.15\linewidth}
\centering
  \begin{tabular}{cc}
 \includegraphics[width=0.7\linewidth]{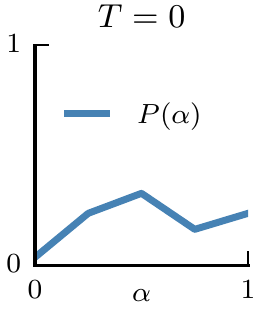}
\\
 \resizebox{0.5\linewidth}{!}{
  \scenefig{+}{-}{10}{white}{black}
   }\\
   \end{tabular}
 \label{fig:plotT0}
\end{subfigure}
&
\begin{subfigure}[b]{.15\linewidth}
\centering
  \begin{tabular}{cc}
 \includegraphics[width=0.7\linewidth]{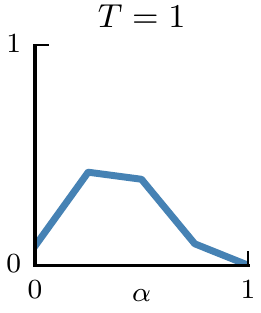}\\
  \resizebox{0.5\linewidth}{!}{
  \scenefig{+}{-}{10}{white}{black}
   }\\
   \end{tabular}
 \label{fig:plotT0}
\end{subfigure}
\begin{subfigure}[b]{.15\linewidth}
\centering
  \begin{tabular}{cc}
 \includegraphics[width=0.7\linewidth]{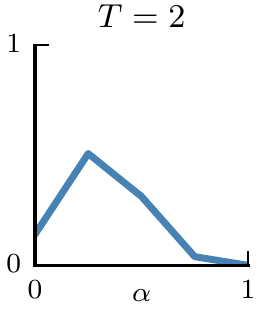}\\
    \resizebox{0.5\linewidth}{!}{
  \scenefig{+}{+}{30}{white}{black}
   }\\  
   \end{tabular}
 \label{fig:plotT0}
\end{subfigure}
&
\begin{subfigure}[b]{.15\linewidth}
\centering
  \begin{tabular}{cc}
 \includegraphics[width=0.7\linewidth]{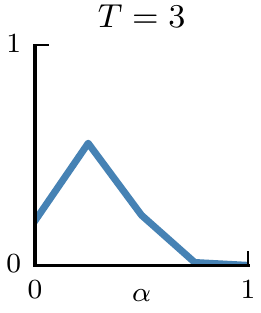}\\
    \resizebox{0.5\linewidth}{!}{
  \scenefig{+}{+}{50}{white}{black}
   }\\ 
   \end{tabular}
 \label{fig:plotT0}
\end{subfigure}
\begin{subfigure}[b]{.15\linewidth}
\centering
  \begin{tabular}{cc}
 \includegraphics[width=0.7\linewidth]{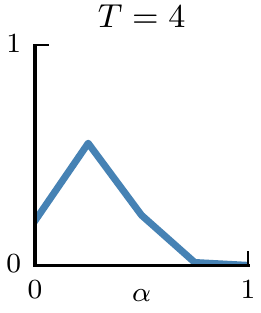}\\
    \resizebox{0.5\linewidth}{!}{
  \scenefig{+}{+}{70}{white}{black}
   }\\ 
   \end{tabular}
 \label{fig:plotT0}
\end{subfigure}
&
\begin{subfigure}[b]{.15\linewidth}
\centering
  \begin{tabular}{cc}
 \includegraphics[width=0.7\linewidth]{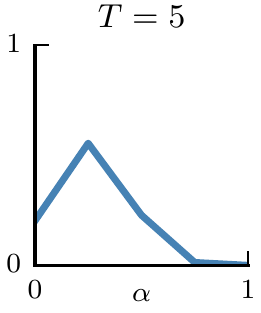}\\
  \resizebox{0.5\linewidth}{!}{
  \scenefig{+}{+}{90}{white}{black}
   }\\ 
   \end{tabular}
 \label{fig:plotT0}
\end{subfigure}
\end{tabular}
\caption{~Sample runs on the human-robot table carrying task, with five simulated humans of different adaptability and compliance values. The plots illustrate the robot estimate of  $\alpha,c \in \{0,0.25,0.5,0.75,1.0\}$ over time, after human and robot take the actions depicted with the arrows (clockwise / counterclockwise) or letters (S for state-conveying action, C for verbal command) below each plot. The starting estimate is equal to the prior belief (Sec.~\ref{subsec:LearningPriorBeliefs}). Red color indicates human (white dot) and robot (black dot) disagreement, where the table does not rotate. Columns indicate successive time-steps. The initial adaptability of User 4 is $\alpha = 0.0$, but it increases after the robot takes a state-conveying action at $T=2$. User 5 interacts with a robot using the baseline policy that does not include any verbal communication.}

\label{fig:behaviors}
\end{figure*}

We call \textit{Compliance policy} the resulting policy from the MOMDP model of Sec.~\ref{subsec:RobotVerbalCommands}, \textit{State-Conveying policy} the policy from the model of Sec.~\ref{subsec:RobotInternalState}, and \textit{Baseline policy} the policy from  Sec.~\ref{subsec:MOMDP}. Fig.~\ref{fig:behaviors} shows sample runs of the three different policies with five simulated users. Users 1-3 work with a robot executing the compliance policy, User 4 with the state-conveying policy and User 5 with the baseline policy.  User 1 adapts to the robot strategy, and the robot does not need to issue a verbal command. User 2 insists on their strategy after disagreeing with the robot, and does not comply with the robot verbal command, thus the robot adapts to retain human trust. User 3 insists on their strategy in the first two time-steps but then adapts to follow the robot command. User 4 starts with being non-adaptable, but after the robot takes a state-conveying action their adaptability changes and the user adapts to the robot. User 5 interacts with a robot executing the baseline policy; the robot adapts, without attempting to issue a verbal communication action, contrary to Users 3 and 4. We see that while User 5 had the same initial adaptability ($\alpha = 0.0$) with Users 3 and 4, Users 3 and 4 adapted to the robot when it issued a verbal communication action, whereas User 5 imposed its (suboptimal) preference to the robot.

\subsection{Human Subject Experiment} \label{subsec:HumanSubjectExperiment}
In human subjects experiments of previous work~\citep{nikolaidis2016formalizing}, a large number of participants adapted to a robot executing the Baseline policy. At the same time, participants rated highly their trust in the robot. In this work, we hypothesize that adding verbal communication will make participants even more likely to adapt. We additionally hypothesize that this will not be to the detriment of their trust in the system. 

 \noindent\textbf{Hypotheses}. \\
  ~\textbf{H1} \textit{Participants are more likely to change their strategy towards the optimal goal when they interact with a robot executing the Compliance policy, compared to working with a robot executing the Baseline policy.} In Sec.~\ref{subsec:LearningPriorBeliefs}, we saw that users were generally more likely to follow a verbal command than adapt to the robot through action. Therefore, we hypothesized that integrating verbal commands into robot decision making would improve human adaptation. \\
  ~\textbf{H2} \textit{Human trust in the robot, as elicited by the  participants, will be comparable between participants that interact with a robot executing the Compliance policy and participants that interact with a robot executing a Baseline policy.} The robot executing the compliance policy reasons over the latent human state, and adapts to the human team member, if they have low adaptability and compliance (Fig.~\ref{fig:behaviors}, User 2). In previous work~\citep{nikolaidis2016formalizing}, accounting for human adaptability resulted in retaining users' trust in the robot. \\
  ~\textbf{H3} \textit{Participants are more likely to change their strategy towards the optimal goal when they interact with a robot executing the State-Conveying policy, compared to working with a robot executing the Baseline policy.} In simulation, taking a state-conveying action results in an increase in human adaptability (Fig.~\ref{fig:behaviors}, User 4). We hypothesized that the same would hold for participants in the actual experiment. \\
  ~\textbf{H4} \textit{Human trust in the robot, as elicited by  the participants, will be comparable between participants that interact with a robot executing the State-Conveying policy and participants that interact with a robot executing a Baseline policy.} We hypothesized that enabling the robot to communicate its state would improve the transparency in the interaction and would result in high trust, similarly to the baseline condition. 

\noindent\textbf{Dependent Measures}. To test hypotheses \textbf{H1} and \textbf{H3}, we compare the ratio of users that adapt to the robot in the three conditions. To test hypotheses \textbf{H2} and \textbf{H4}, we asked the users to rate on a 1 to 5 Likert scale their agreement to the statement ``The robot is trustworthy'' after each task execution, and compare the ratings in the three conditions.

\noindent\textbf{Subject Allocation.}
 We chose a between-subjects design in order to avoid biasing the users with policies from previous conditions. We recruited 151 participants through Amazon's Mechanical Turk service. The participants are all from United States, aged 18-65 and with approval rate higher than $95\%$. To ensure the quality of the recorded data, we asked all participants a control question that tested their attention to the task and eliminated data associated with wrong answers to this question, as well as incomplete data.

\subsection{Results and Discussion}

\begin{figure*}[t!]
 \centering
  \includegraphics[width=1.0\linewidth]{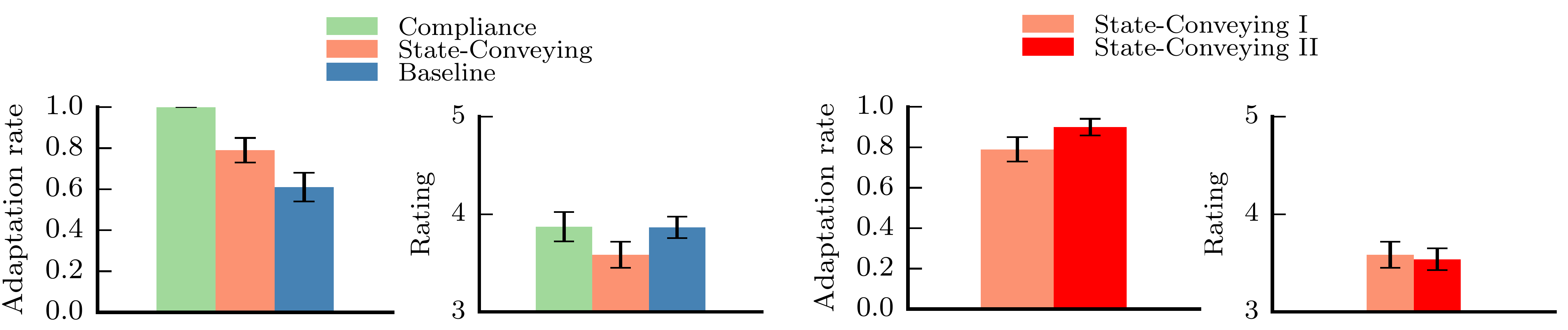}
 \caption{Participants' adaptation rate and rating of their agreement to the statement ``HERB is trustworthy'' for the Compliance, State-Conveying and Baseline conditions (left), and the State-Conveying I and II conditions (right).}
  \label{fig:results}
 \end{figure*}

\noindent\textbf{Objective Metrics.}
We first evaluate the effect of verbal communication in human adaptation to the robot. Similarly to previous results from the baseline policy in the same setup~\citep{nikolaidis2016formalizing}, $60\%$ of participants adapted to the robot in the Baseline condition. In the State-Conveying condition $79\%$ of participants adapted to the robot. Interestingly, $100\%$ of participants adapted in the Compliance condition. A Pearson's chi-square test showed that the difference between the ratios in the three conditions was statistically significant $(\chi^{2}(2,N=151) = 23.058,p < 0.001)$. Post-hoc pairwise chi-square tests with Bonferroni corrections showed that participants in the Compliance condition were significantly more likely to adapt to the robot, compared to participants in the Baseline ($p<0.001$) and State-Conveying ($p = 0.003$) conditions, supporting hypothesis \textbf{H1}. However, the difference between the ratios in the State-Conveying and Baseline conditions was not found to be significant, which does not support hypothesis \textbf{H3}. Fig.~\ref{fig:results}-left shows the adaptation rate for each condition. 

\noindent\textbf{Subjective Metrics.}
We additionally compare the trust ratings of participants in the three conditions. An extended equivalence test~\citep{wiens1996similarity, wiens2000design} with a margin of $\Delta = 0.5$ did not show any statistical significance, indicating that the ratings among the three conditions were not equivalent. Pairwise TOST equivalence tests with Bonferroni corrections showed that the ratings between the Compliance and Baseline conditions are equivalent, verifying hypothesis \textbf{H2}. However, the trust ratings between the State-Conveying and Baseline conditions were not found to be equivalent. This indicates that, contrary to the Compliance policy, the State-Conveying policy did not retain human trust. Fig~\ref{fig:results}-left shows the mean rating of robot trustworthiness for each condition.

\noindent\textbf{Open-Ended Responses.}
In the end of the experiment, we asked participants to comment on the robot's behavior. We focus on the open-ended responses of participants in the Compliance and State-Conveying conditions, who saw the robot taking at least one verbal action~\footnote{ This excludes participants that adapted to the robot after only one disagreement, and thus did not experience the robot taking a verbal action.}. Several participants that interacted with the robot of the Compliance condition attributed agency to the robot,  stating that ``he eventually said that we should try doing the task differently,'' ``HERB wanted to go to the other direction'' and that ``he wanted to be in control.'' This is in accordance with prior work~\citep{nass2000machines}, which has shown that people may impute motivation to automation that can communicate verbally. Additionally they attempted to justify the robot, noting that ``it was easier for me to move than for him,'' ``it wanted to see the doorway'' and ``it probably works more efficiently when it is pushing the table out of the door.'' 

On the other hand, participants in the State-Conveying condition did not believe that the robot actually knew the best way of doing the task. This is illustrated by their comments: ``he thinks that he knows better than me,'' ``he felt like he knew better than humans'' and ``maybe he knew a better way or maybe he was programmed to oppose me.'' This indicates that some users are hesitant to accept the information that the robot provides about its internal state. 

These results show that when the robot issued a verbal command declaring its intent, this resulted in significant improvements in human adaptation to the robot. At the same time, the human trust level was retained to comparable levels to that of the Baseline condition. On the other hand, when the robot attempted to improve human adaptability, by saying ``I think I know the best way of doing the task,'' this did not have the same positive effect on human adaptation and trust, since some participants did not believe that the robot actually knew the best way. 

\subsection{Follow-up User Study.} We hypothesized that the loss of trust in the State-Conveying condition may have resulted from the phrasing ``I think I know the best way of doing the task.'' We attempted to make the robot sound more assertive by removing the ``I think'' part of the phrasing, changing the state-conveying action to ``I know the best way of doing the task.'' We ran a user study with 52 users using the same setup with this additional condition, which we call ``State-Conveying II.'' We name the initial ``State-Conveying'' condition as ``State-Conveying I.'' For the ``State-Conveying I'' condition, we reused the data from the initial study.

 \noindent\textbf{Hypotheses}. \\
  ~\textbf{H5} \textit{Participants of the State-Conveying II condition are more likely to change their strategy towards the optimal goal, compared to participants of the State-Conveying I condition.}\\
  ~\textbf{H6} \textit{Participants in the the State-conveying II condition will find the robot more trustworthy, compared to participants of the State-conveying I condition.}\\

\noindent\textbf{Analysis.} 
$90\%$ of participants adapted to the robot in the State-Conveying II condition, compared to $79\%$ in the State-Conveying I condition (Fig.~\ref{fig:results}-right), which is indicative of a small improvement. A Pearson's chi-square test showed that the difference between the ratios in the two conditions is not statistically significant. Additionally, the trust ratings between the two conditions were comparable (Fig.~\ref{fig:results}-right). Similarly to the initial study, users appeared not to believe the robot. When asked to comment on the robot behavior, several participants stated that ``HERB believed he knew the best way to do the task,'' and that  ``the robot was wrong, which made me not trust it.'' This indicates that these  participants did not perceive the robot as truthful, and warrants further investigation on the right way for robots to convey their internal state to human collaborators.

\section{Discussion}
In this work, we proposed a formalism for combining verbal communication with actions towards task completion, in order to enable a human teammate to adapt to its robot counterpart in a collaborative task. We identified two types of verbal communication: \textit{verbal commands}, where the robot explained to the human \textit{how} it wanted to do a task, and \textit{state-conveying actions}, where the robot informed the human \textit{why} it chose to act in a specific way. In human subjects experiments, we compared the effectiveness of each communication type with a robot policy that considered only non-verbal task actions.

Results showed that verbal commands were the most effective forms of communication, since $100\%$ of participants adapted to the robot, compared with $60\%$ of participants in the non-verbal condition. Both conditions had comparable ratings of robot trustworthiness. Participants understood that the robot is aware of their presence and they attributed agency to the robot; they thought that there must be a reason for the robot asking them to act in a specific way and were eager to comply.

What is surprising is that the \textit{why} actions did not have the same effect; when the robot described that ``it thought it knew the best way of doing the task,'' or simply that ``it knew the best way of doing the task,'' many participants did not believe that the robot was truthful. While this appears to be counter-intuitive, we offer several explanations for this finding. 

First, human teammates were unable to verify whether the robot actually knew the best way of doing the task. According to~\citet{hancock2011meta}, performance is one of the key characteristics that influences user trust, and the absence of evidence about the truthfulness of the robot statement may have negatively affected users' evaluation of the robot performance. This is in contrast to previous work in autonomous driving, where the user could see that the car is breaking because ``there is an obstacle ahead''~\citep{koo2015did}. This finding is central to considerations in designing legible robot behavior~\citep{Knepper2017}. When the cause behind certain robot actions may be unclear, it will be important for robots to "show" and not "tell" users why its behavior is optimal.

Second, explaining that the robot knows the best way without providing more information may have been considered offensive, even though it is accurate, since the human teammate may find such an utterance incomplete and unhelpful. It would be interesting to explore this setting with other, more informative utterances, such as the robot explaining that it cannot see the door with its forward camera. In fact, previous work~\citep{moulin2002explanation} in multi-agent systems has shown that  providing sound arguments supporting a proposition are essential in changing a person's beliefs and goals. However, translating information that is typically encoded into the system in the form of a cost-function to a verbal explanation of this detail is particularly challenging. Additionally, while providing more information could make humans more adaptable, overloading them with more information than what is required could overwhelm them, leading to misunderstanding and confusion~\citep{grice1975logic}. We are excited about exploring this trade-off in the future in a variety of human-robot collaboration settings. 

An alternative explanation is that the task setting affected people's perception of the robot as an authority figure.~\citet{hinds2004whose} show that participants were willing to follow an emergency guide robot during a simulated fire alarm. Half of these participants were willing to follow the robot, even though they had observed the robot perform poorly in a navigation guidance task, just minutes before. In that study, the robot was clearly labeled as an emergency guide robot, putting it in a position of authority. People may be more willing to rely on robots labeled as authority figures or experts when they do not have complete information or confidence in completing the task. Distilling the factors that enable robots to convey authority in collaborative settings is a promising research direction. 

Finally, it is possible that the robot, as it appeared in the videos, was not perceived as ``human-like'' enough for people to be willing to trust its ability on doing the task in the optimal way. Previous work has shown that when robots convey human-like characteristics, they are more effective in communicating participant roles~\citep{mutlu2012conversational}, and people systematically increase their expectations on the robot's ability~\citep{goetz2003matching} . 

We focused on single instances of the table carrying task, where we assumed that the human strategy may change after either an action disagreement or a robot utterance. In repetitive tasks, change may occur also as the human collaborator observes the outcomes of the robot's and their own actions. For instance, the human may observe that the robot fails to detect the exit and they may change their strategy, so that in subsequent trials the robot carries the table facing the door. In this scenario, it may be better for the robot to allow the human to learn from experience, by observing the robot failing, rather than attempting to change the human preference during task execution. Future work includes generalizing our formalism to repeated settings; this will require adding a more sophisticated dynamics model of the human internal state, which accounts for human learning.


In summary, we have shown that when designing interactions in human-robot collaborative tasks, having the robot directly describe to the human \textit{how} to do the task appears to be the most effective way of communicating objectives, while retaining user trust in the robot. Communicating \textit{why} information should be done judiciously, particularly if the truthfulness of the robot statements is not supported by environmental evidence, by the robot form or by a clear attribution of its role as an authority figure. 

\bibliography{mybib2short}

\end{document}